\documentclass[conference]{IEEEtran}

\usepackage{amsmath}
\usepackage[]{algorithm2e}
\usepackage{algpseudocode}
\usepackage{flushend}

\begin{document}

\title{Reliable Evaluation of Neural Network for Multiclass Classification of Real-world Data}

\author{\IEEEauthorblockN{Siddharth Dinesh}
\IEEEauthorblockA{Department of Computer Science and Information Systems\\
Birla Institute of Technology and Science Pilani\\
K.K. Birla Goa Campus, Goa 403726, India\\
E-mail: f2012519@goa.bits-pilani.ac.in}
\and
\IEEEauthorblockN{Tirtharaj Dash}
\IEEEauthorblockA{Department of Computer Science and Information Systems\\
Birla Institute of Technology and Science Pilani\\
K.K. Birla Goa Campus, Goa 403726, India\\
E-mail: tirtharaj@goa.bits-pilani.ac.in}}

\maketitle

\begin{abstract}
This paper presents a systematic evaluation of Neural Network (NN) for classification of real-world data. In the field of machine learning, it is often seen that a single parameter that is `predictive accuracy' is being used for evaluating the performance of a classifier model. However, this parameter might not be considered reliable given a dataset with very high level of skewness. To demonstrate such behavior, seven different types of datasets have been used to evaluate a Multilayer Perceptron (MLP) using twelve(12) different parameters which include micro- and macro-level estimation. In the present study, the most common problem of prediction called `multiclass' classification has been considered. The results that are obtained for different parameters for each of the dataset could demonstrate interesting findings to support the usability of these set of performance evaluation parameters.
\end{abstract}

\section{Introduction}
Machine Learning (ML) has been a well-explored domain of research in the present decade. It is basically a method of data analysis that automatically builds models from historical data. ML uses algorithms that iteratively learn from such historical data, which in turn finds hidden insights and patterns inside the data without being explicitly programmed for the it. ML techniques have been employed in many different real world problems such as fraud detection, intrusion detection, web search engines, e-mail spam filtering, sentiment analysis, credit scoring, equipment failures prediction, pattern and image recognition, genetics and genomics, robotics (see \cite{1,2,3,4}). For all such real world tasks, ML requires certain algorithms which are called `learning algorithms' based on which the pattern inside the historical data could be explored. Based on definition of the problem and availability of data, a learning algorithm could be of two major types: supervised, or unsupervised. Supervised learning trains a model in the presence of a supervisor (technically an `error' term) whereas the unsupervised algorithms do not require such an error term for training. Supervised learning methods could be applied to cases of prediction and classification. This paper is more focused in the area of supervised ML.

Supervised ML allows access to the data labels during training and testing phases of the model. For example, there could be a problem of identifying how a student will perform in the present semester given his attendance, weekly performance, participation in the class, past records. Such a problem requires huge historical student records and their performance. Supervised ML tries to understand such characteristic in the data and predicts the performance of the presently considered student. Consider a set of data records $\mathcal{D}_1,\mathcal{D}_2,\ldots,\mathcal{D}_n$ that have to be assigned to a set of predefined labels, or classes $c_1,c_2,\ldots,c_m$ ($m$ is usually less than $n$). The process of assigning a class label to a data record is termed as classification. Classification falls into two major categories such as binary classification, multiclass classification. Binary classification is one of the basic classification task where $m$ is 2. For example, finding out whether the performance of the student would be `good' or `bad'; here, `good' and `bad' are two categorical classes. However, classification of a data record where more than two classes are available (i.e. $m>2$) is one of the challenging tasks and is called multiclass classification. For example, finding out whether the performance of the student would be `excellent', `good', `average', `below average', or     `poor'. Evaluating a learning model for binary classification is easier as compared to evaluation of a learning model for multiclass classification because of the following reason. In the binary classification problem, a data record can be identified as true positive ($tp$) or true negative ($tn$) or false positive ($fp$) or false negative ($fn$). Generation of all these information during testing could be presented as a matrix called `confusion matrix' (Table \ref{tab:confusionmat}). The performance measures which could be used to evaluate a learning model for binary classification are accuracy, sensitivity (recall), specificity, precision, $F-$score. These parameters could be computed using Equations \eqref{eq:acc} through to Equation \eqref{eq:fscore}. For more information on these parameters, one can see the published work of Sokolova and Lapalme \cite{5}. $Accuracy$ gives an overall estimate of predictive power of a model. $Pricision$ gives information about positive predictive value of the model. $Sensitivity$ and $Specificity$ estimate the true positive rate and true negative rate for the testing dataset. $F-score$ is a balanced mean between $Precision$ and $Sensitivity$.

\begin{table}[h]
\centering
\caption{Confusion matrix for binary classification}
\setlength{\tabcolsep}{0.5em}
\bgroup
\def\arraystretch{1.5}
\begin{tabular}{|c|c|c|}
\hline 
Data Class & Classified as $pos$ & Classified as $neg$ \\ 
\hline 
$pos$ & $tp$ & $fn$ \\ 
\hline 
$neg$ & $fp$ & $tn$ \\ 
\hline 
\end{tabular} 
\egroup
\label{tab:confusionmat}
\end{table}

\begin{equation}
\label{eq:acc}
Accuracy=\frac{tp+tn}{tp+fn+fp+tn}
\end{equation}
\begin{equation}
\label{eq:sen}
Sensitivity (recall)=\frac{tp}{tp+fn}
\end{equation}
\begin{equation}
\label{eq:spe}
Speicificity=\frac{tn}{fp+tn}
\end{equation}
\begin{equation}
\label{eq:prec}
Precision=\frac{tp}{tp+fp}
\end{equation}
\begin{equation}
\label{eq:fscore}
F-score=\frac{(\beta^2+1)tp}{(\beta^2+1)tp+\beta^2 fn+fp}
\end{equation}

The above-mentioned performance measures are helpful for evaluating classifiers in binary classification problems. However, researchers often use accuracy for evaluating the performance of classifiers for multiclass classification tasks. This is because of the fact that the evaluation of the above performance measures are quite difficult when more than two classes are there to be considered for the problem (e.g. multiclass classification of student performance in the semester). Moreover, accuracy measure for multiclass classification task could be non-reliable when the dataset (a set of data records) is skewed more towards a particular class. For such kind of problems, a reliable evaluation of a classifier would be very much crucial. In this work, we limit our discussion to popularly used supervised learning model (classifier) called \textit{Neural Network (NN)}\footnote{Also known as Artificial Neural Network (ANN)}. We focus our attention majorly on reliable performance evaluation of NN for real-world multiclass classification problems rather than reasoning about the obtained results. We estimate various performance measures which could be used for multiclass classification problems based on the information provided in \cite{5}.

The rest of the paper consists of following sections. Section \ref{sec:2} provides an explanation of the implemented NN and its training algorithm. Section \ref{sec:3} details on the tested benchmark datasets used in this work. Section \ref{sec:4} expands on the results and discussion followed by conclusion in section \ref{sec:5}.

\section{NN and its Training with Gradient Descent}
\label{sec:2}
NN is a biologically inspired mathematical model which is used to approximate functions that depend on a set of inputs called `features'. Computational processing of NN closely follows information processing inside the human brain which has a complex network of neurons. The motivation behind evaluating NN in this work is that they have high adaptation power given a better learning algorithm and their rigorous applications in many different real-life problems. Moreover, there is a good amount of flexibility to tune a learning algorithm with different parameters to improve the performance of the NN. Generally, NN is a layered architecture where neurons (nodes) are arranged in layers. In this work, multilayered NN, specifically, a multilayer perceptron (MLP), has been implemented. MLP architecture comprises of an input layer, a hidden layer, and an output layer. The input layer takes values of the features from the dataset and computes an output in the output layer. The hidden layer is responsible for transforming the input features into a set of features which could be processed by the output layer. It has been seen that the performance of the NN depends on the learning algorithm based on which it has been trained using the training dataset. The on-line learning has been made popular by researchers by the development of back-propagation algorithm that uses gradient descent strategy for minimization of error during training of the NN. It should be noted that the stochastic version of the gradient descent algorithm, called stochastic gradient descent (SGD), has been used in this work where the weights of the NN are updated for each random sample from the training dataset. An NN can be trained effectively by back-propagation if a sufficiently large dataset is used during training. The training dataset refers to a set of data records with known class (i.e. the class to which the data record belongs). Let a data instance be represented as a pair of vectors $\left(\vec{x},\vec{t}\right)$, where $\vec{x}$ is the vector of input features, and $\vec{t}$ is the vector of target  output values or classes. Let us denote $i$th record in the dataset as $\vec{x_i}=\left(x_{i_1}, x_{i_2}, x_{i_3},\ldots,x_{i_{n_{in}}}\right)$ and the class labels for this data record as a set $\{t_i\}$ (see footnote\footnote{$t_i$ is a set because a class can be represented as a set of multiple classes e.g. If there are three classes of data, then a data belonging to class 1, class 2 and class 3 could be represented as \{0,0,1\}, \{0,1,0\} and \{0,0,1\} respectively.}). So, the $i$th training data instance in a training dataset can be represented as $(x_{i_1}, x_{i_2}, x_{i_3}, \dots, x_{i_{n_{in}}},\{t_i\})$, where $n_{in}$ is the number of inputs to the neural net. Let the number of neurons in the hidden layer be represented as $n_{hidden}$ and number of neurons in the output layer as $n_{out}$. Each neuron in one layer of the NN is connected to each neuron in its next layer with a weight value, which represents the strength of the connection. We denote these weight set as a vector $\overrightarrow{W}=\{w_{ij}\}$, where $1 \leq i,j \leq \left(n_{in}+1\right)n_{hidden}+\left(n_{hidden}+1\right)n_{out}$. $\overrightarrow{W}$ also includes a set of biases in the hidden layer and the output layer. The weights including the biases of an NN is called as the knowledge base (KB) of the NN. The KB of the NN is updated during training of the NN. The back-propagation based training algorithm \cite{6} of the NN has been presently briefly in Algorithm \ref{alg:backprop} followed by a set of governing equations.

\begin{algorithm}[h]
\caption{Training of NN with back-propagation}
\label{alg:backprop}
Initialize the weights $\overrightarrow{W}$ to small random numbers\;
\While{Stopping criteria not met}{
\For{each training pattern $\left(\vec{x_i},\vec{t_i}\right)$}{
Process the input forward using Eq. \eqref{eq:forward}\;
Propagate the error backward through the network using equations Eq. \eqref{eq:delk}--Eq. \eqref{eq:wupd}\;
}
}
\end{algorithm}

During the forward processing, the output of NN can be obtained by multiplying weights and the input pattern instance as shown in Eq. \eqref{eq:forward}.
\begin{equation}
\label{eq:forward}
o=f\left(\overrightarrow{W}.\vec{x}\right)
\end{equation}
where, $f$ is the activation function of the output unit and is usually a sigmoid function as given in Eq. \eqref{eq:sigmoid}.
\begin{equation}
\label{eq:sigmoid}
\sigma(y)=\frac{1}{1+e^{-c.y}}
\end{equation}
where, $c$ is a non-negative constant and is set to 1 in this work.

The back-propagation training algorithm attempts to minimize an error term, $\delta$ (for supervised classification tasks) by changing or updating the weights of the NN. The error term is basically the squared error between the net output values and the target values for the corresponding input instance (Refer Eq. \eqref{eq:err}). Please note that when $t_i$ is represented as a set of ones and zeros, Eq. \eqref{eq:err} might not be suitable. In this work, as batch learning is used, it uses a modified version of the following equation. The details have been provided in the section \ref{sec:4}.
\begin{equation}
\label{eq:err}
E(\overrightarrow{W})=\frac{1}{2}\sum_{i}^{}{\sum_{k\in outputs}^{}{(t^k_i-o^k_i)^2}}
\end{equation}
where, $outputs$ is the set of output units in the NN; $t^k_i$ and $o^k_i$ are the target and output values associated with the $k$th output unit for the $i$th input instance.

For each network output unit $k$, the error term $\delta_k$ can be computed using Eq. \eqref{eq:delk}.
\begin{equation}
\label{eq:delk}
\delta_k=o_k(1-o_k)(t_k-o_k)
\end{equation}
Similarly, for each hidden unit $h$, the error term $\delta_h$ can be computed using Eq. \eqref{eq:delh}. 
\begin{equation}
\label{eq:delh}
\delta_h=o_h(1-o_h)\sum_{k\in outputs}^{}{w_{kh}\delta_k}
\end{equation}

The weight update equation for the network are as given in Eq. \eqref{eq:wupd} where, $x_{j_i}$ is the value of $j$th feature of the $i$th data record.
\begin{equation}
\label{eq:wupd}
w_{ji}=w_{ji}+\Delta w_{ji}
\end{equation}
and,
\begin{equation}
\label{eq:delw}
\Delta w_{ji}=\eta \delta_jx_{j_i}
\end{equation}
In the Eq. \eqref{eq:delw}, the constant $\eta$ is the learning rate.
It should be noted that, we also used a constant term in the weight update rule called `momentum factor' (denoted as $\mu$) to the weight update rule, which makes the amount of weight update on the $n$th training iteration depend partially on the weight update that had occurred during the $(n-1)$th training iteration, which can be clearly understood from Eq. \eqref{eq:momentum}.
\begin{equation}
\label{eq:momentum}
\Delta w_{ji}(n)=\eta \delta_j x_{j_i}+\mu\Delta w_{ji}(n-1)
\end{equation}
Here $\Delta w_{ji}(n)$ is the weight update performed during the $n$th iteration; the constant $\mu$ is usually fixed in the range $[0,1)$ prior to training. In our work, the learning rate, $\eta$ and the momentum factor, $\mu$ are set to 0.3 and 0.1 respectively.

\section{Datasets}
\label{sec:3}
To evaluate the performance of multiclass classification problem with NN, following seven different real-world benchmark datasets have been used in this work. All the datasets have been obtained from UCI Machine Learning repository \cite{7}. However, we briefly explain all the datasets with regard to their dimension. More details about each dataset could be obtained from \cite{7}.
\subsubsection*{Abalone dataset}
The goal of using Abalone dataset is to predict abalone age through the number of rings on the shell given various descriptive attributes of the abalone. There are 4177 data instance each with 8 input features and a class label.
\subsubsection*{Breast Cancer dataset}
This dataset is one of the popular medical benchmarks in ML research. The patient records have been obtained from University of Wisconsin Hospital, Madison. There are total 699 records with 10 input features (one is an ID which is not used for computation) and a class label.
\subsubsection*{E-coli dataset}
This dataset contains protein localization sites of 336 proteins with 8 input features (one is a sequence number and not being used for computation) and class label.
\subsubsection*{Glass dataset}
This dataset is used to identify and predict the type of glass based on 9 different manufacturing features. It has 214 records containing such information.
\subsubsection*{ILPD dataset}
ILPD refers to Indian Liver Patient Database contains, 583 liver patient records each has been marked as a liver patient or a non-liver patient as their type.
\subsubsection*{Iris dataset}
Iris dataset is a very well known benchmark dataset which contains 4 input features and a class attribute. The dataset
contains total 150 instances, 50 of each type of plants such as Iris Setosa, Iris Versicolour, and Iris Virginica.
\subsubsection*{Wine}
It contains the chemical analysis results of wines grown in Italy. There are 13 predicting features and a class attribute. There are total 178 instances.

\section{Performance Evaluation}
\label{sec:4}
All the simulations in this work are carried out in MATLAB R2015b using a personal computer system with Windows 10 operating system, quad-core processor with the equal clock rate of 1.70 GHz and main memory of 4 GB.

\subsection{Preparation of data for simulation}
It is important to prepare data wisely before training of the NN. The real world data obtained from UCI ML repository are distributed non-uniformly and hence, they can not be used directly during training and testing of the NN. Therefore, the input features were initially normalized in the range [0,1]. The normalized dataset was then partitioned into training and an independent test set in the ratio 70:30. The process of training and testing have been repeated for 10 independent runs (simulations) to get the average performance of the NN and its performance deviation from the mean.
\subsection{Performance measures}
The performance parameters which are evaluated for multiclass classification are described as follows. For a class $c_i$, the classifier performance can be assessed with $tp_i$, $fn_i$, $tn_i$ and $fp_i$ and can be calculated from counts of testing instance belonging to $c_i$. The quality of the overall classification performance can be assessed in two different ways such as micro and macro averaging. Macro averaging treats all the classes equally while micro averaging favors classes with more data instances. Computation of various performance measures suitable for evaluating NN for multiclass classification problem can be obtained as follows which is a generalization of the parameters presented in Table \ref{tab:confusionmat} for many classes $c_i$ \cite{5}. For a class $c_i$, $tp_i$, $fn_i$, $tn_i$ and $fp_i$ counts respectively. Micro- and macro-averaging indices are represented by $\mu$ and $M$ respectively.

Average accuracy ($Accuracy$) could be used to evaluate average per-class effectiveness of the NN and can be computed as,
\begin{equation}
\label{eq:aacc}
Accuracy=\frac{\sum_{i=1}^{m}{\frac{tp_i+tn_i}{tp_i+fn_i+fp_i+tn_i}}}{m},
\end{equation}
where, $m$ is the number of classes.

Other crucial measures could be obtained from Equation \eqref{eq:mprec} through to Equation \eqref{eq:Mfscore} \cite{5}.
\begin{equation}
\label{eq:mprec}
Precision_{\mu}=\frac{\sum_{i=1}^{m}{tp_i}}{\sum_{i=1}^{m}{(tp_i+fp_i)}}
\end{equation}
\begin{equation}
\label{eq:Mprec}
Precision_M=\frac{\sum_{i=1}^{m}{\frac{tp_i}{tp_i+fp_i}}}{m}
\end{equation}
\begin{equation}
\label{eq:mspec}
Specificity_{\mu}=\frac{\sum_{i=1}^{m}{tn_i}}{\sum_{i=1}^{m}{(fp_i+tn_i)}}
\end{equation}
\begin{equation}
\label{eq:Mspec}
Specificity_M=\frac{\sum_{i=1}^{m}{\frac{tn_i}{fp_i+tn_i}}}{m}
\end{equation}
\begin{equation}
\label{eq:msens}
Sensitivity_{\mu}(recall_{\mu})=\frac{\sum_{i=1}^{m}{tp_i}}{\sum_{i=1}^{m}{(tp_i+fn_i)}}
\end{equation}
\begin{equation}
\label{eq:Msens}
Sensitivity_M(recall_M)=\frac{\sum_{i=1}^{m}{\frac{tp_i}{tp_i+fn_i}}}{m}
\end{equation}
\begin{equation}
\label{eq:mfscore}
F-score_{\mu}=\frac{(\beta^2+1)Precision_{\mu}Recall_{\mu}}{\beta^2Precision_{\mu}Recall_{\mu}}
\end{equation}
\begin{equation}
\label{eq:Mfscore}
F-score_{\mu}=\frac{(\beta^2+1)Precision_MRecall_M}{\beta^2Precision_MRecall_M}
\end{equation}

Apart from the above mentioned micro and macro measures, the training error (which is mean-squared error during training, $MSE_{train}$), testing error (which is mean-squared error during training, $MSE_{test}$), and the training time ($Time_{train}$) have also been noted in this work. However, it is wise to mention that the SGD does not require $MSE$ during training rather it requires the error term ($\delta$) between a random sample and its prediction for updating the weights of the NN. $MSE$ has been computed as parameter to observe the average error of convergence for the model during training and testing. $MSE$ can be calculated using the following equation,
\begin{equation}
\label{eq:mse}
MSE=\frac{1}{n}\sum_{i=1}^{n}{(t_i-o_i)},
\end{equation}
where, $n$ is the number of data records considered during the process (either training or testing), $t_i$ is the hypothesis or the target for $i$th data instance, and $o_i$ is the output of the NN for the $i$th data instance.

\subsection{Results}
The number of neurons in the hidden layer ($n_{hidden}$) is one of the most important architectural parameters which directly influences performance of NN during training and capturing data for preparing a knowledge base. However, the setting of this parameter apriori has been an unsolved problem in ML research \cite{3}. In this work, $n_{hidden}$ has been set to 60, 80 and 100 and the results have been noted for each of the datasets. All the obtained results have been summarized and depicted as tables for different $n_{hidden}$ values. Table \ref{tab:res_60} depicts results obtained for $n_{hidden}=60$. Similarly, Table \ref{tab:res_80} and Table \ref{tab:res_100} present results obtained for $n_{hidden}=80$ and $n_{hidden}=100$ respectively. It should be noted that all the results shown in these three tables are averaged over ten(10) independent simulations for each of the datasets.

\begin{table*}[]
\centering
\caption{Performance of NN with $n_{hidden}=60$ averaged over 10 independent runs (Datasets are presented in the columns; first column lists the performance measures; `--' means `could not be obtained')}
\label{tab:res_60}
\setlength{\tabcolsep}{0.5em}
\bgroup
\def\arraystretch{1.3}
\begin{tabular}{|c|c|c|c|c|c|c|c|}
\hline
\textbf{Measures} & \textbf{Abalone} & \textbf{Breast Cancer} & \textbf{E-coli} & \textbf{Glass} & \textbf{ILPD} & \textbf{Iris} & \textbf{Wine} \\ \hline
\textbf{$MSE_{train}$} & 0.7985$\pm$0.0017 & 0.0276$\pm$0.0064 & 0.1281$\pm$0.0157 & 0.2808$\pm$0.0256 & 0.3265$\pm$0.0138 & 0.0293$\pm$0.0126 & 0.0004$\pm$0.0001 \\ \hline
\textbf{$Time_{train}$} & 432.9480$\pm$16.8208 & 36.1701$\pm$0.1070 & 20.5751$\pm$0.8514 & 12.7597$\pm$0.0353 & 31.4476$\pm$0.1180 & 6.2847$\pm$0.0304 & 11.9831$\pm$0.4870 \\ \hline
\textbf{$MSE_{test}$} & 1.0503$\pm$0.2312 & 0.6359$\pm$0.1888 & 0.9739$\pm$0.1709 & 0.9532$\pm$0.0915 & 0.4994$\pm$0.0563 & 0.4693$\pm$0.1446 & 0.3680$\pm$0.0987 \\ \hline
\textbf{$Accuracy$} & 0.9241$\pm$0.0047 & 0.6388$\pm$0.1266 & 0.8337$\pm$0.0343 & 0.7693$\pm$0.0322 & 0.6690$\pm$0.1009 & 0.7941$\pm$0.1494 & 0.8151$\pm$0.0394 \\ \hline
\textbf{$Precision_{\mu}$} & 0.1100$\pm$0.0411 & 0.6388$\pm$0.1266 & 0.4380$\pm$0.0660 & 0.3078$\pm$0.0966 & 0.6690$\pm$0.1009 & 0.6911$\pm$0.2241 & 0.7226$\pm$0.0590 \\ \hline
\textbf{$Precision_M$} & -- & 0.7398$\pm$0.1208 & -- & -- & 0.4910$\pm$0.1042 & 0.8903$\pm$0.0495 & 0.8087$\pm$0.0472 \\ \hline
\textbf{$Specificity_{\mu}$} & 0.9604$\pm$0.0025 & 0.6388$\pm$0.1266 & 0.9019$\pm$0.0233 & 0.8616$\pm$0.0193 & 0.6690$\pm$0.1009 & 0.8456$\pm$0.1120 & 0.8613$\pm$0.0295 \\ \hline
\textbf{$Specificity_M$} & 0.9575$\pm$0.0020 & 0.5767$\pm$0.0420 & 0.8578$\pm$0.0281 & 0.8329$\pm$0.0156 & 0.5057$\pm$0.0115 & 0.8415$\pm$0.1128 & 0.8457$\pm$0.0270 \\ \hline
\textbf{$Sensitivity_{\mu}$} & 0.1100$\pm$0.0411 & 0.6388$\pm$0.1266 & 0.4380$\pm$0.0660 & 0.3078$\pm$0.0966 & 0.6690$\pm$0.1009 & 0.6911$\pm$0.2241 & 0.7226$\pm$0.0590 \\ \hline
\textbf{$Sensitivity_M$} & 0.0445$\pm$0.0016 & 0.5767$\pm$0.0420 & 0.1583$\pm$0.0513 & 0.1952$\pm$0.0745 & 0.5057$\pm$0.0115 & 0.6904$\pm$0.2250 & 0.6539$\pm$0.0262 \\ \hline
\textbf{$F-score_{\mu}$} & 0.1100$\pm$0.0411 & 0.6388$\pm$0.1266 & 0.4380$\pm$0.0660 & 0.3078$\pm$0.0966 & 0.6690$\pm$0.1009 & 0.6911$\pm$0.2241 & 0.7226$\pm$0.0590 \\ \hline
\textbf{$F-score_M$} & -- & 0.6460$\pm$0.0715 & -- & -- & 0.4953$\pm$0.0627 & 0.8581$\pm$0.0857 & 0.7358$\pm$0.0360 \\ \hline
\end{tabular}
\egroup
\end{table*}

\begin{table*}[]
\centering
\caption{Performance of NN with $n_{hidden}=80$ averaged over 10 independent runs (Datasets are presented in the columns; first column lists the performance measures; `--' means `could not be obtained')}
\label{tab:res_80}
\setlength{\tabcolsep}{0.5em}
\bgroup
\def\arraystretch{1.3}
\begin{tabular}{|c|c|c|c|c|c|c|c|}
\hline
\textbf{Measures} & \textbf{Abalone} & \textbf{Breast Cancer} & \textbf{E-coli} & \textbf{Glass} & \textbf{ILPD} & \textbf{Iris} & \textbf{Wine} \\ \hline
\textbf{$MSE_{train}$} & 0.7976$\pm$0.0030 & 0.0300$\pm$0.0070 & 0.1224$\pm$0.0171 & 0.3200$\pm$0.0290 & 0.3258$\pm$0.0141 & 0.0251$\pm$0.0155 & 0.0004$\pm$0.0001 \\ \hline
\textbf{$Time_{train}$} & 584.5762$\pm$0.9668 & 47.3517$\pm$0.4763 & 25.7683$\pm$0.1250 & 23.8334$\pm$3.8817 & 40.8900$\pm$0.157 & 8.1097$\pm$0.0234 & 15.2530$\pm$0.3440 \\ \hline
\textbf{$MSE_{test}$} & 1.1077$\pm$0.2381 & 0.5045$\pm$0.1037 & 0.8640$\pm$0.2212 & 0.8134$\pm$0.1116 & 0.4656$\pm$0.0638 & 0.3117$\pm$0.1765 & 0.2041$\pm$0.1062 \\ \hline
\textbf{$Accuracy$} & 0.9246$\pm$0.0049 & 0.7057$\pm$0.0921 & 0.8468$\pm$0.0221 & 0.7901$\pm$0.0349 & 0.7161$\pm$0.0372 & 0.9289$\pm$0.0658 & 0.9270$\pm$0.0530 \\ \hline
\textbf{$Precision_{\mu}$} & 0.1151$\pm$0.0484 & 0.7057$\pm$0.0921 & 0.4680$\pm$0.0480 & 0.3906$\pm$0.1169 & 0.7161$\pm$0.0372 & 0.8933$\pm$0.0987 & 0.8906$\pm$0.0795 \\ \hline
\textbf{$Precision_M$} & -- & 0.8009$\pm$0.0669 & -- & -- & 0.6609$\pm$0.1811 & 0.9298$\pm$0.0617 & 0.9139$\pm$0.0618 \\ \hline
\textbf{$Specificity_{\mu}$} & 0.9606$\pm$0.0026 & 0.7057$\pm$0.0921 & 0.9104$\pm$0.0139 & 0.8730$\pm$0.0209 & 0.7161$\pm$0.0372 & 0.9467$\pm$0.0494 & 0.9453$\pm$0.0397 \\ \hline
\textbf{$Specificity_M$} & 0.9574$\pm$0.0020 & 0.6184$\pm$0.0549 & 0.8694$\pm$0.0267 & 0.8574$\pm$0.0169 & 0.5150$\pm$0.0197 & 0.9471$\pm$0.0497 & 0.9424$\pm$0.0413 \\ \hline
\textbf{$Sensitivity_{\mu}$} & 0.1151$\pm$0.0484 & 0.7057$\pm$0.0921 & 0.4680$\pm$0.0480 & 0.3906$\pm$0.1169 & 0.7161$\pm$0.0372 & 0.8933$\pm$0.0987 & 0.8906$\pm$0.0795 \\ \hline
\textbf{$Sensitivity_M$} & 0.0438$\pm$0.0021 & 0.6184$\pm$0.0549 & 0.1698$\pm$0.0339 & 0.3562$\pm$0.1276 & 0.5150$\pm$0.0197 & 0.8852$\pm$0.1184 & 0.8871$\pm$0.0820 \\ \hline
\textbf{$F-score_{\mu}$} & 0.1151$\pm$0.0484 & 0.7057$\pm$0.0921 & 0.4680$\pm$0.0480 & 0.3906$\pm$0.1169 & 0.7161$\pm$0.0372 & 0.8933$\pm$0.0987 & 0.8906$\pm$0.0795 \\ \hline
\textbf{$F-score_M$} & -- & 0.6963$\pm$0.0515 & -- & -- & 0.5712$\pm$0.0780 & 0.9191$\pm$0.0788 & 0.9000$\pm$0.0714 \\ \hline
\end{tabular}
\egroup
\end{table*}

\begin{table*}[]
\centering
\caption{Performance of NN with $n_{hidden}=100$ averaged over 10 independent runs (Datasets are presented in the columns; first column lists the performance measures; `--' means `could not be obtained')}
\label{tab:res_100}
\setlength{\tabcolsep}{0.5em}
\bgroup
\def\arraystretch{1.3}
\begin{tabular}{|c|c|c|c|c|c|c|c|}
\hline
\textbf{Measures} & \textbf{Abalone} & \textbf{Breast Cancer} & \textbf{E-coli} & \textbf{Glass} & \textbf{ILPD} & \textbf{Iris} & \textbf{Wine} \\ \hline
\textbf{$MSE_{train}$} & 0.7983$\pm$0.0018 & 0.0303$\pm$0.0056 & 0.0078$\pm$0.0059 & 0.0555$\pm$0.0097 & 0.3366$\pm$0.0120 & 0.1228$\pm$0.0179 & 0.0004$\pm$0.0001 \\ \hline
\textbf{$Time_{train}$} & 675.5284$\pm$13.6632 & 58.1381$\pm$0.1665 & 108.1947$\pm$0.2115 & 52.1537$\pm$1.7554 & 50.5680$\pm$0.1651 & 96.1866$\pm$0.2467 & 18.7710$\pm$0.3396 \\ \hline
\textbf{$MSE_{test}$} & 1.0478$\pm$0.1976 & 0.4228$\pm$0.1350 & 0.5167$\pm$0.1967 & 0.6140$\pm$0.1898 & 0.4190$\pm$0.0452 & 0.6902$\pm$0.1833 & 0.1386$\pm$0.0685 \\ \hline
\textbf{$Accuracy$} & 0.9242$\pm$0.0060 & 0.7541$\pm$0.0941 & 0.6746$\pm$0.1393 & 0.8777$\pm$0.0326 & 0.7080$\pm$0.0366 & 0.6224$\pm$0.0994 & 0.9509$\pm$0.0240 \\ \hline
\textbf{$Precision_{\mu}$} & 0.1198$\pm$0.0462 & 0.7541$\pm$0.0941 & 0.6746$\pm$0.1393 & 0.5660$\pm$0.0981 & 0.7080$\pm$0.0366 & 0.6224$\pm$0.0994 & 0.9264$\pm$0.0361 \\ \hline
\textbf{$Precision_M$} & -- & 0.8440$\pm$0.0532 & 0.7682$\pm$0.1062 & -- & 0.5828$\pm$0.0724 & 0.5487$\pm$0.0280 & 0.9358$\pm$0.0305 \\ \hline
\textbf{$Specificity_{\mu}$} & 0.9604$\pm$0.0032 & 0.7541$\pm$0.0941 & 0.6746$\pm$0.1393 & 0.9288$\pm$0.0196 & 0.7080$\pm$0.0366 & 0.6224$\pm$0.0994 & 0.9632$\pm$0.0180 \\ \hline
\textbf{$Specificity_M$} & 0.9572$\pm$0.0020 & 0.6821$\pm$0.0705 & 0.6751$\pm$0.0528 & 0.9084$\pm$0.0328 & 0.5218$\pm$0.0197 & 0.5354$\pm$0.0312 & 0.9615$\pm$0.0201 \\ \hline
\textbf{$Sensitivity_{\mu}$} & 0.1198$\pm$0.0462 & 0.7541$\pm$0.0941 & 0.6746$\pm$0.1393 & 0.5660$\pm$0.0981 & 0.7080$\pm$0.0366 & 0.6224$\pm$0.0994 & 0.9264$\pm$0.0361 \\ \hline
\textbf{$Sensitivity_M$} & 0.0460$\pm$0.0022 & 0.6821$\pm$0.0705 & 0.6751$\pm$0.0528 & 0.2432$\pm$0.0512 & 0.5218$\pm$0.0197 & 0.5354$\pm$0.0312 & 0.9295$\pm$0.0377 \\ \hline
\textbf{$F-score_{\mu}$} & 0.1198$\pm$0.0462 & 0.7541$\pm$0.0941 & 0.6746$\pm$0.1393 & 0.5660$\pm$0.0981 & 0.7080$\pm$0.0366 & 0.6224$\pm$0.0994 & 0.9264$\pm$0.0361 \\ \hline
\textbf{$F-score_M$} & -- & 0.7533$\pm$0.0579 & 0.7178$\pm$0.0748 & -- & 0.5492$\pm$0.0395 & 0.5418$\pm$0.0276 & 0.9326$\pm$0.0331 \\ \hline
\end{tabular}
\egroup
\end{table*}

\subsection{Discussion}
Discussion regarding the obtained results (as depicted in Table \ref{tab:res_60}, Table \ref{tab:res_80} and Table \ref{tab:res_100}) in this work has been primarily based on various performance parameters rather than how the values are obtained. This work summarizes different performance parameters to use for evaluation of NN or similar classifier for multiclass classification of real-world data. It has been seen that the popularly known `accuracy' parameter could not be as a reliable parameter for proper evaluation of a classifier. Hence, in this work, diversified datasets are being used for evaluation of the NN for the classification problem.

The training performances of the NN have been approximately equal for the Abalone dataset with different $n_{hidden}$ settings such as $n_{hidden}=60$, $n_{hidden}=80$ and $n_{hidden}=100$. Moreover, the standard deviation in the $MSE_{train}$ is very low for all the three cases. However, it is obvious that increasing the number of the hidden neurons increases the architectural complexity of the NN and hence the training time ($Time_{train}$). Moreover, when the $n_{hidden}$ is set to as high as 100, there might be a high probability of over-fitting training data and could not achieve the better performance than performances observed for other considered settings such as $n_{hidden}=60$ or $n_{hidden}=80$. It is also evident from the obtained average classification accuracy. Moreover, given such a high accuracy of approximately 92\%, the positive predictive value that is $Precision$ does not seem to be satisfactory. Similarly, all other parameters such as $Sensitivity$, $F-score$ are not reliable. However, the true negative rate ($Specificity$) is quite good for all the tree different settings.

The performance results which have been obtained for the Breast Cancer dataset is quite better than that of Abalone. With the increase in the number of hidden neurons, the NN is able to predict the hypothesis for the test dataset with high accuracy. The error of convergence decreases with increase in $n_{hidden}$ which could be possible by capturing the features properly during the training of the NN. The performance parameters such as $Accuracy$, $Precision_\mu$, $Precision_M$, and other mentioned parameters seem to be following similar property. Hence, it could be assumed that predictive accuracy could be a good measure for this dataset. However, $Accuracy$ parameter alone could not be taken as a reliable parameter while evaluating NN during multiclass classification of the Abalone dataset.

The evaluated performance of the NN for E-coli dataset closely follows the discussion about the performance for the Abalone dataset. Unlike the Abalone dataset, the predictive accuracy of the NN for the E-coli dataset dropped quite high with $n_{hidden}=100$. Similar cases could also be seen for other evaluated parameters for the same dataset. The performance of NN for the Glass dataset is comparatively improved with the increase in $n_{hidden}$ along with other performance parameters. However, although the predictive accuracy is good, the micro- and macro-sensitivity seems to be compromised even though the input feature is transformed into a higher dimensional feature set by the increase in $n_{hidden}$. Therefore, for Glass dataset, predictive accuracy might not be a suitable parameter for evaluation of NN.

Evaluation of the NN with regard to ILPD dataset is quite good as compared to recent literature (see \cite{8}) considering the fact that the accuracy and other parameters still follow a particular limit of deviation unlike the results obtained for Abalone, E-coli and Glass datasets. This means that the micro and macro parameters could be considered reliable given such a low accuracy for the dataset. The performance of the NN for Iris classification is superior when $n_{hidden}=80$ as compared to the results obtained for other two settings. The positive predictive rate and negative predictive rates are also better as compared to those with $n_{hidden}=60$ and $n_{hidden}=100$. However, assuming that the predictive accuracy values are low for these two mentioned settings, it could be seen that the other performance evaluation parameters still reveal reliable performance for this dataset. However, not much deviation over the results obtained for the parameters could be seen for the Wine dataset where the performance for the $Accuracy$ and other micro and macro parameters are improving with the increase in $n_{hidden}$. Moreover, given the present setup of experimentation, one could also achieve slightly different results because of the fact that the initial weights and biases of the NN are fixed at random. If a proper weight set is fixed initially, one could possibly land up in obtaining a better results for the same settings. This argument could be supported by the fact that the gradient descent may not always guarantee a close-to-optimal weight set at the end of the NN training process.

\section{Conclusion}
\label{sec:5}
In this work, a detailed evaluation of NN classifier has been carried out for multiclass classification of real-world data. It has been seen that the use of predictive accuracy as a single parameter for evaluating an NN would not be wise given high skewness in the test data. Results obtained for different types of datasets clearly show that although the accuracy is very high, there could be fair chance that the positive or negative predictive rate falls far below any reliable range. In this work, this type of property has been seen in the performance of NN for a majority of tested datasets such as Abalone dataset, E-coli dataset, Glass dataset. Hence, it would be wise to use many different parameters for such classification problems to accurately evaluate a classifier.

\end{document}